\documentclass[3p,times,fleqn]{elsarticle}
\usepackage{amsmath}
\usepackage{amssymb}
\usepackage{color}
\usepackage{stfloats}
\usepackage{mathrsfs}
\usepackage{tabularx}
\usepackage{booktabs}
\usepackage{colortbl}
\usepackage{rotating}
\usepackage{boldline}
\usepackage{float}
\usepackage{bm}
\usepackage{changes}
\usepackage{gensymb}
\usepackage{hyperref}
\DeclareMathOperator*{\argmin}{arg\,min}
\DeclareMathOperator*{\median}{\textrm{median}}
\newcommand{\R}{\mathbb{R}}

\usepackage{todonotes}
\setuptodonotes{size=\small}

\begin{document}
\renewcommand{\topfraction}{0.98}	
\renewcommand{\bottomfraction}{0.98}
\setcounter{topnumber}{3}
\setcounter{bottomnumber}{3}
\setcounter{totalnumber}{4}         
\setcounter{dbltopnumber}{4}        
\renewcommand{\dbltopfraction}{0.98}	
\renewcommand{\textfraction}{0.05}	
\renewcommand{\floatpagefraction}{0.5}	
\renewcommand{\dblfloatpagefraction}{0.5}	
\newcommand{\beq}{\begin{equation}}
\newcommand{\eeq}{\end{equation}}
\newcommand{\divg}{\mbox{\rm{div}}\,}
\newcommand{\Divg}{\mbox{\rm{Div}}\,}
\newcommand{\D}  {\displaystyle}
\newcommand{\DS} {\displaystyle}
\newcommand{\RM}[1]{\textit{\MakeUppercase{\romannumeral #1{}}}}
\newtheorem{remark}{\bf{{Remark}}}
\def\sca   #1{\mbox{\rm{#1}}{}}
\def\mat   #1{\mbox{\bf #1}{}}
\def\vec   #1{\mbox{\boldmath $#1$}{}}
\def\scas  #1{\mbox{{\scriptsize{${\rm{#1}}$}}}{}}
\def\scaf  #1{\mbox{{\tiny{${\rm{#1}}$}}}{}}
\def\vecs  #1{\mbox{\boldmath{\scriptsize{$#1$}}}{}}
\def\tens  #1{\mbox{\boldmath{\scriptsize{$#1$}}}{}}
\def\tenf  #1{\mbox{{\sffamily{\bfseries {#1}}}}}
\def\ten   #1{\mbox{\boldmath $#1$}{}}
\def\Ass  {\overset{\hspace*{0.4cm} n_{\scas{el}}}
          {\underset{\scaf{c},\scaf{d}=1}{\msf{A}}}}
\def\ltr   #1{\mbox{\sf{#1}}}
\def\bltr  #1{\mbox{\sffamily{\bfseries{{#1}}}}}
\sloppy
\begin{frontmatter}
\title{\Large Ensemble learning of the atrial fiber orientation with physics-informed neural networks}

\author[01]{Efraín Magaña}
\author[04,05]{Simone Pezzuto}
\author[01,02,03]{Francisco~Sahli Costabal}

%
\address[01]{Department of Mechanical and Metallurgical Engineering, School of Engineering, Pontificia Universidad Cat\'olica de Chile, Santiago, Chile}
\address[04]{Laboratory of Mathematics for Biology and Medicine, Department of Mathematics, University of Trento, Italy}

\address[05]{Euler Institute, Università della Svizzera italiana, Switzerland}
\address[02]{Institute for Biological and Medical Engineering, Schools of Engineering, Medicine and Biological Sciences, Pontificia Universidad Cat\'olica de Chile, Santiago, Chile}

\address[03]{Millennium Institute for Intelligent Healthcare Engineering, iHEALTH}


\begin{abstract} %

The anisotropic structure of the myocardium is a key determinant of the cardiac function. To date, there is no imaging modality to assess in-vivo the cardiac fiber structure. We recently proposed Fibernet, a method for the automatic identification of the anisotropic conduction---and thus fibers---in the atria from local electrical recordings. Fibernet uses cardiac activation as recorded during electroanatomical mappings to infer local conduction properties using physics-informed neural networks. In this work, we extend Fibernet to cope with the uncertainty in the estimated fiber field. Specifically, we use an ensemble of neural networks to produce multiple samples, all fitting the observed data, and compute posterior statistics. We also introduce a methodology to select the best fiber orientation members and define the input of the neural networks directly on the atrial surface. With these improvements, we outperform the previous methodology in terms of fiber orientation error in 8 different atrial anatomies. Currently, our approach can estimate the fiber orientation and conduction velocities in under 7 minutes with quantified uncertainty, which opens the door to its application in clinical practice. We hope the proposed methodology will enable further personalization of cardiac digital twins for precision medicine.
\end{abstract}
\end{frontmatter}


\section{Introduction}\label{intro}


Cardiac arrhythmias are diseases at epidemic scale, with a worldwide prevalence more than 37 million cases only for atrial fibrillation (AF). The prevalence of AF has increased by 33\,\% during the last 20 years, and it is expected to grow even larger in the next years, especially in middle-income countries~\cite{AHA2020,lippi2021global}. The clinical treatment of AF also remains suboptimal. Standard therapies like cardiac ablation of the pulmonary veins are successful in restoring the normal rhythm only in 50\,\% to 70\,\% of the patients \cite{baykaner2023mapping}. In patients with persistent or permanent AF, the disease is highly dependent on the structural and functional properties of the atrial tissue, which is affected by scars~\cite{Dutta2013} and fibrosis~\cite{Schotten2011}.

Cardiac myocytes are organized in the form of fibers. The orientation of these fibers changes throughout the heart, with some consistency between individuals \cite{streeter1969fiber}. The fiber structure plays a major role in the electromechanical function of the heart, since the active force is mainly exerted in the fiber direction~\cite{roberts1979influence}. The electrical wave that drives the mechanical contraction is propagated faster along the fiber orientation.
Despite the prominent role of the fiber orientation, its characterization \textit{in vivo} remains an open challenge. Studies from \textit{ex vivo} histology have shaped the current knowledge that we have about fiber orientation \cite{ho2009importance}. Recently, diffusion tensor imaging (DT Imaging) has become an alternative to determine the fiber orientation in beating hearts \cite{Toussaint2010MEDIA}. This is a magnetic resonance imaging protocol that has been used in static organs, such as the brain, for many years. However, the application to the heart while it is moving is a recent development \cite{Stoeck2016MRM, Edelman1994MRM}.
In less than a decade from its inception, DT Imaging has already been linked to diseases in the ventricles \cite{Gotschy2021JACC, Das2023JACC, VonDeuster2016Circ, Ferreira2014JCMR}. 
The current use of this technique is limited to the ventricles, due to the poor image resolution and long acquisition times.  The usage of this method in the atria has only been \textit{ex vivo} \cite{pashakhanloo2016myofiber} where there is no movement, as the atrial wall is much thinner than the ventricular wall. 

The role of remodeling of cardiac fibers in AF is less understood. The lack of fiber imaging methods has lead to a lack of understanding of the link between the atrial fiber orientation and diseases.
Recent methods have been developed to infer the atrial fiber orientation from electrical measurements \cite{roney2019,GranditsPINN2021,Ruiz2022}. Here, the entire conduction velocity tensor is inferred, and the fiber orientation corresponds to the direction of fastest propagation. The input data required can be recorded minimally-invasively with an electrode that registers the arrival times of the electrical wave at different locations inside the atrial chamber, constructing the so-called electroanatomical maps. These methods also obtain the conduction velocities, which are critical for the reliable simulation of the electrophysiological behavior of the atria.  Patient-specific models that incorporate the learned fibers and conduction velocities can be used to test the efficacy of different ablation treatments \textit{in silico}, which can later be performed by physicians, potentially reducing recurrence rates of AF \cite{boyle2019computationally}. 

Recovering the fiber orientation from electrical measurements is an ill-posed inverse problem. An emerging technique for this kind of problem is physics-informed neural networks~\cite{raissi2019physics}. In this paradigm, neural networks are used to represent the quantities to be inferred, and the networks are trained towards 2 objectives: learn the data available and also satisfy the physics of the problem, typically expressed in the form of a partial differential equation. This technique has been successfully applied to problems in cardiac electrophysiology~\cite{herrero2022ep,costabal2020eikonal}. In this work, we extend our previous methodology to infer the atrial fiber orientation with physics-informed neural networks~\cite{Ruiz2022} from electroanatomical maps to improve its robustness, accuracy and clinical applicability. In particular, we propose 3 improvements: first, we create positional encoding that is defined directly on the atrial surface rather than cartesian coordinates of the space where the surface is embedded. This change in input for the physics-informed neural networks creates an improvement in accuracy for the fiber orientation reconstruction. Second, we provide uncertainty estimates by training an ensemble of neural networks. This estimates can be used in clinical practice to obtain a measurement of the trustworthiness of the models, and also can be used to guide the acquisition of more data. Third, we develop a novel way to select the most accurate fiber orientation from ensemble members, which significantly improves the accuracy of our method. Finally, we completely refactored our implementation, leading to a fiber orientation inference in less than 7 minutes with quantified uncertainty, making the application of this method to clinical practice closer. 

This paper is organized as follows: Section~\ref{sec:methods} introduces the forward model of cardiac electrophysiology, the methodology to infer the fiber direction with physics-informed neural networks, how we perform the uncertainty quantification and the methods to select the best fiber orientation. In Section~\ref{sec:numerical}, we test the proposed algorithm in 8 different atrial anatomies and assess the effect of noise and data density. We also compare the different fiber selection method. We end the paper with a discussion in Section 4.


\begin{figure}[h!tbp]
    \centering
    \includegraphics[width=0.9\linewidth]{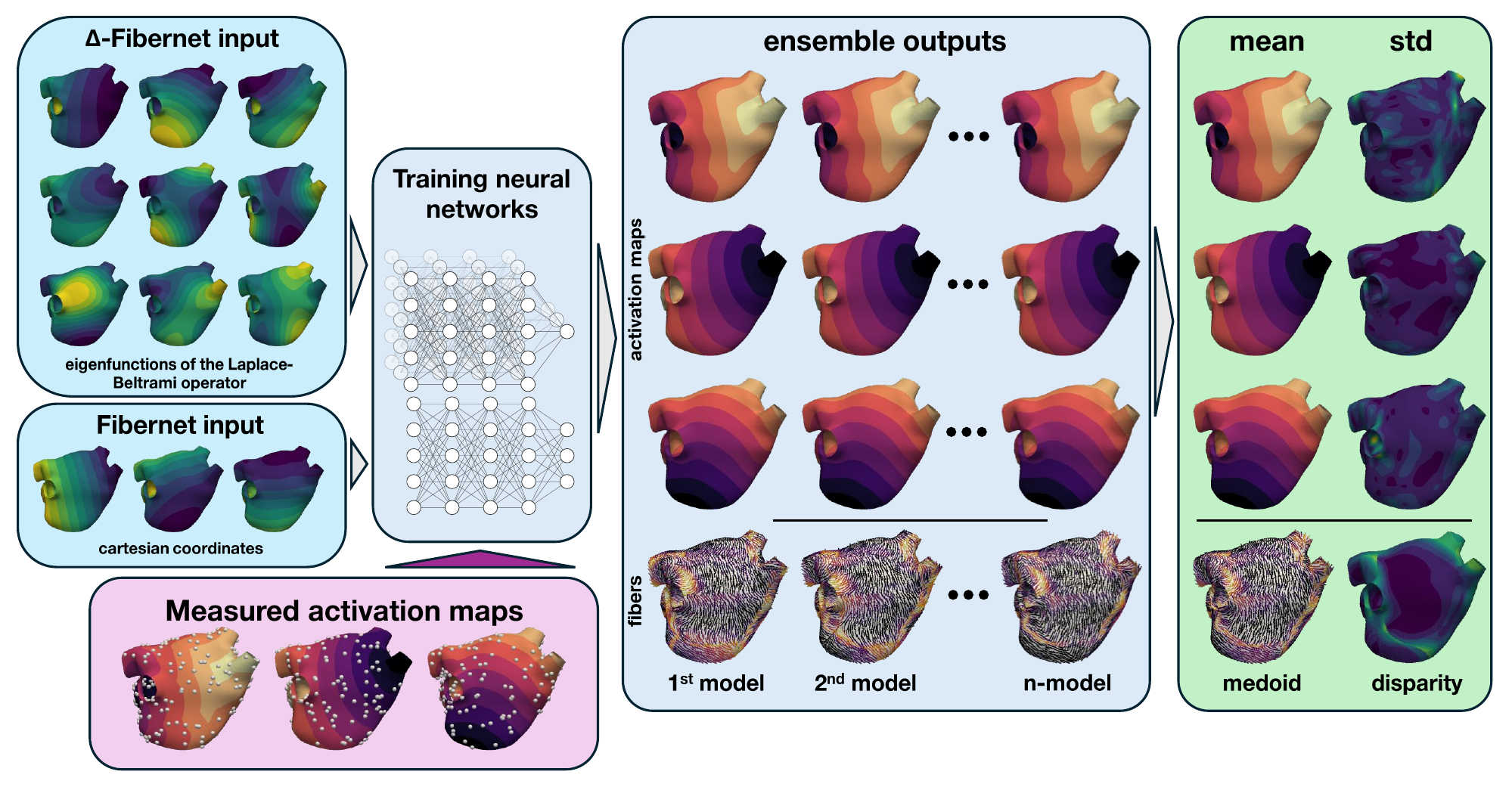}
    \caption{Schematic highlighting the difference between the input of $\Delta$-Fibernet and Fibernet, and the uncertainty quantification of the ensemble results.}
    \label{fig:diagram}
\end{figure}
\section{Methods}
\label{sec:methods}

In this section we begin by reviewing the forward model of arrival times in cardiac electrophysiology, then the inverse problem to recover the fibers, finalizing with the proposed methods for uncertainty quantification.

\subsection{Propagation model}
The electrical activity of the heart can be accurately modeled by the bidomain or monodomain equations, which results in a detailed representation of the spatiotemporal dynamics of the transmembrane potential~\cite{franzone2014mathematical}. However, the electrical wave propagation can also be modeled only by its arrival time at different locations, removing the temporal dimension of the problem~\cite{Pezzuto2017,GranditsGEASI2021}. To do this, the arrival times $\phi(\vec{x})$, also known as activation times, are modeled through the anisotropic eikonal equation:
\begin{equation}
    \sqrt{\ten{D}(\vec{x}) \nabla \phi(\vec{x}) \cdot \nabla \phi(\vec{x})} =1\, \quad \vec{x}\in \mathcal{S}.
    \label{eq:aniso}
\end{equation}
To simplify the problem, we assume that the atrial wall is thin enough to be represented as a surface $\mathcal{S} \in \R^3$.
$\ten{D}(\vec{x})$  is a symmetric, positive-definite tensor field representing the conduction velocities of the electrical wave. We define it as,
\begin{equation}
\ten{D}(\vec{x}) = v_l(\vec{x})^2 \vec{l}(\vec{x}) \otimes \vec{l}(\vec{x}) + v_t(\vec{x})^2 \vec{t}(\vec{x}) \otimes \vec{t}(\vec{x})+ v_n(\vec{x})^2 \vec{n}(\vec{x}) \otimes \vec{n}(\vec{x}),\end{equation}
where $\vec{l}(\vec{x}) \in \R^3$ correspond to the longitudinal direction, that is the direction of the cardiac fibers, $\vec{t}(\vec{x}) \in \R^3$ corresponds to the transverse direction, and $\vec{n}(\vec{x}) \in \R^3$ the normal direction of the surface. As the wave does not propagate in the normal direction, we assume $v_n(\vec{x})=0, \forall \vec{x} \in \mathcal{S}$.
We define the fiber and transverse directions on a tangent space to the surface, defined by the orthonormal basis $\mathcal{B}(\vec{x}) = \{p_1(\vec{x}), p_2(\vec{x})\}$ as
\begin{align}
    \vec{l}(\vec{x}) &= \cos(\alpha(\vec{x}) \vec{p}_1(\vec{x}) \otimes \vec{p}_1(\vec{x}) +\sin(\alpha(\vec{x})) \vec{p}_2(\vec{x}) \otimes \vec{p}_2(\vec{x}),\\
    \vec{t}(\vec{x}) &= -\sin(\alpha(\vec{x})) \vec{p}_1(\vec{x}) \otimes \vec{p}_1(\vec{x}) +\cos(\alpha(\vec{x})) \vec{p}_2(\vec{x}) \otimes \vec{p}_2(\vec{x}),
\end{align}
where $\alpha(\vec{x})$ correspond to the angle between the fiber direction and $\vec{p}_1(\vec{x})$. This defines the tensor $\ten{D}$ on the basis $\mathcal{B}$ as $\ten{D}_\mathcal{B}$. Then we use the projection to global coordinates to use Eq.~\eqref{eq:aniso}. These steps are summarized in matrix form as follows:
\begin{equation}
    \ten{D}_{\mathcal{B}}(\vec{x}) = 
  \begin{bmatrix}
    \cos(\alpha(\vec{x})) &  -\sin(\alpha(\vec{x}))&0\\
    \sin(\alpha(\vec{x}))& \cos(\alpha(\vec{x})) & 0\\
    0& 0& 1
  \end{bmatrix}
  \begin{bmatrix}
    v_l(\vec{x})^2 &  0&0\\
    0& v_t(\vec{x})^2 & 0\\
    0& 0& 0
  \end{bmatrix}
  \begin{bmatrix}
    \cos(\alpha(\vec{x})) &  \sin(\alpha(\vec{x}))&0\\
    -\sin(\alpha(\vec{x}))& \cos(\alpha(\vec{x})) & 0\\
    0& 0& 1
  \end{bmatrix},
\end{equation}
\begin{equation}
    \ten{D}(\vec{x}) = 
  \begin{bmatrix}
    \vrule  &  \vrule & \vrule\\
    \vec{p}_1(\vec{x})& \vec{p}_2(\vec{x}) & \vec{n}(\vec{x})\\
    \vrule& \vrule& \vrule
  \end{bmatrix} D_\mathcal{B}(\vec{x})
  \begin{bmatrix}
    \text{---} & \vec{p}_1(\vec{x}) & \text{---}\\
    \text{---} & \vec{p}_2(\vec{x}) & \text{---}\\
    \text{---} & \vec{n}(\vec{x}) & \text{---}
  \end{bmatrix}
\end{equation}

Following these definitions, we will work with a parameterized version of the tensor $\ten{D}(\vec{x})$:
\begin{equation}
    \ten{D}(\vec{x}) = \tilde{\ten{D}}(\ten{d}(\vec{x})),
\end{equation}
where 
\begin{equation}
    \ten{d}(\vec{x}) = [a(\vec{x}), e_1(\vec{x}), e_2(\vec{x})],
\end{equation}
and $a(\vec{x}):=\cos\left(\alpha\left(\vec{x}\right)\right) \in [-1,1]$, the cosine of the angle of the fiber on the basis $\mathcal{B}$, $e_1(\vec{x}):=v_l(\vec{x})^2 >0$ is the square of the longitudinal velocity, and $e_2(\vec{x}):=v_t(\vec{x})^2 >0$ is the square of the transverse velocity.


\subsection{Learning fibers from multiple maps}

With this propagation model, it is possible to set up an inverse problem to determine a field fiber orientation, given by $\alpha(\vec{x})$ and the conduction velocities $v_l(\vec{x})$, $v_t(\vec{x})$ from measurements of the activation times $\phi(\vec{x})$. This data is acquired clinically in electroanatomical maps. We have previously shown that data coming from at least 3 different pacing sites is required to properly reconstruct these quantities \cite{Ruiz2022}. 
In general, we will consider the problem of simultaneously identifying $\ten{D}(\vec{x})$ and $\phi_i$ with $i \in [1,N]$, $N$ being the number of activation maps. Specifically, we extend the PINN framework developed in \cite{Ruiz2022}. Here, we approximate each map $\phi_i$ and the conduction velocity vector parameters $\vec{d}$ with an independent neural network, such that:
\begin{align}
    \ten{d}(\vec{x}) & \approx \hat{\vec{d}}(\vec{x})\coloneqq NN(\vec{x}; \vec{\theta}_D),&\label{eq:nnd}\\
    \phi_i(\vec{x})&\approx \hat \phi_i(\vec{x})\coloneqq NN(\vec{x}; \vec{\theta}_{\phi_i})& i\in[1,N]\label{eq:nnphi}.
\end{align}
Optimizing the parameters of the neural networks $(\vec{\theta}_D , \vec{\theta}_{\phi_i})$ via the loss function defined as:
\begin{equation}
    \mathcal{L}(\vec{\theta}_D , \vec{\theta}_{\phi})\coloneqq \lambda_{\text{data}}\mathcal{L}_\text{data}(\vec{\theta}_{\phi})+\lambda_{\text{eiko}}\mathcal{L}_\text{eiko}(\vec{\theta}_{D},\vec{\theta}_{\phi})+\lambda_{\text{CV}}\mathcal{L}_\text{CV}(\vec{\theta}_{D})+\lambda_{\text{ang}}\mathcal{L}_\text{ang}(\vec{\theta}_{D})
\end{equation}
where $\vec{\theta}_{\phi}$ denotes all parameters $\vec{\theta}_{\phi_i}$. The $\mathcal{L}_\text{data}$ term encourages the network to learn the acquired measurements of activations times $\phi$. The term $\mathcal{L}_\text{eiko}$ drives to network to satisfy the eikonal equation~\eqref{eq:aniso} and links the activation times networks and the conduction velocity network. The terms $\mathcal{L}_\text{CV}$ and $\mathcal{L}_\text{ang}$ regularize the conduction velocities and the fiber direction, respectively. The hyperparameters $\lambda_{\text{data}}$, $\lambda_{\text{eiko}}$, $\lambda_{\text{CV}}$ and $\lambda_{\text{ang}}$ modulate the importance of the different loss components. More specifically, we define each loss term as follows:
\begin{align}
    \mathcal{L}_{\text{data}}\coloneqq&\frac{1}{N}\sum_{i=1}^N\frac{1}{N_i}\sum_{k=1}^{N_i}\left(\hat \phi_i(\vec{x_k})-\phi_{i,k}\right)^2,\\
    \mathcal{L}_{\text{eiko}}\coloneqq&\frac{1}{N N_C}\sum_{i=1}^N\sum_{j=1}^{N_C}\left(T_{i,max}\sqrt{\ten{\tilde D}(\vec{\hat d}(\vec{x_j})) \nabla \hat \phi_i(\vec{x_j})\cdot\nabla \hat \phi_i(\vec{x_j})} -1\right)^2,\\
    \mathcal{L}_{\text{CV}}\coloneqq&\frac{1}{N_C}\sum_{j=1}^{N_C}\left( H_{\delta_e}(\nabla\hat e_1(\vec{x_j}))+H_{\delta_e}(\nabla\hat e_2(\vec{x_j}))\right),\\
    \mathcal{L}_{\text{ang}}\coloneqq&\frac{1}{N_C}\sum_{j=1}^{N_C}\left( H_{\delta_a}(\nabla\hat a(\vec{x_j}))\right),
\end{align}
where $N_i$ correspond to the number of measurements on the map $\phi_i$ with $T_{i,max}$ being the maximum activation time registered. The term $\mathcal{L}_\text{eiko}$ will be enforced in $N_C$ collocation points. For regularization of the conduction velocity tensor, we use the Huber Total Variation, which combines the $L_1$ and $L_2$ norms of the gradient: 
\begin{equation}
    H_\delta(\vec{q}) = \begin{cases}
        \frac{1}{2\delta}\|\vec{q}\|^2, \quad \text{for} \quad \|\vec{q}\|\leq\delta\\
        \|\vec{q}\|-\frac{\delta}{2}, \quad \text{otherwise.}
    \end{cases}
\end{equation}

In this work, we extend the methodology presented in~\cite{Ruiz2022} by first acknowledging that we are representing the atria as a surface embedded in $R^3$, which presents a unique set of challenges for learning. In this regard, we follow the methodology proposed in \cite{Sahli2024} to redefine the input of the neural networks to a space that is better suited for learning in surfaces. We replace the input coordinates of the neural networks $\vec{x}$ for the $N_e$ eigenfunctions $v(\vec{x})$ of the Laplace-Beltrami operator of the mesh that describes the surface, see Figure~\ref{fig:diagram} for a graphical representation of the change of input coordinates. One of the main drawbacks of representing the atrial surface with its cartesian coordinates is that 2 points that are close in this space, might be far in a geodesic sense. Using the eigenfunctions of the Laplace-Beltrami operator as inputs alleviates this problem, as distances in this space can resemble geodesic distances~\cite{lipman2010biharmonic}. Now, we modify equations (\ref{eq:nnd}) and (\ref{eq:nnphi}) to the following:

\begin{align}
    &\hat{\ten{D}}(\vec{x})\coloneqq NN\left(\vec{v}(\vec{x}); \vec{\theta}_D\right),&\\
    &\hat \phi_i(\vec{x})\coloneqq NN\left(\vec{v}(\vec{x}); \vec{\theta}_{\phi_i}\right)& i\in[1,N].
\end{align}
We will refer to the original formulation that uses cartesian coordinates as Fibernet and the model that uses the Laplace-Beltrami eigenfunctions as $\Delta$-Fibernet. We approximate the Laplace-Beltrami eigenfunctions with the finite element method. The drawback of this approach is that the dependency of the neural networks on $\vec{x}$ is lost, and automatic differentiation cannot be used to compute the gradient ($\nabla_{\vec{x}}[\cdot]$), required for the loss terms $\mathcal{L}_\text{eiko}$, $\mathcal{L}_\text{CV}$ and $\mathcal{L}_\text{ang}$. For this reason, we also use the finite element method to approximate the gradient. For a scalar field $u$,  we have that,
\begin{equation}
    \nabla_x u \approx \ten{R}^e \ten{B}^e \cdot \left[ \begin{array}{l}
        u_1  \\
        u_2 \\
        u_3
    \end{array}\right],
\end{equation}
where $\ten{B}^e$ correspond to the gradient operator on the local coordinates of the triangular element, $\ten{R}^e$ the rotation matrix of the local triangular element coordinates to the global coordinates and $[u_1, u_2, u_3]^T$ correspond to the scalar field evaluated on each of the nodes of the element. The explicit formulation can be seen on the repository of this paper.

\subsection{Uncertainty quantification}
Given that we are solving an ill-posed inverse problem, there are multiple conduction velocity tensors that may fit the activation maps measured. There are even more network parameters $\vec{\theta}_D, \vec{\theta}_\phi$ that explain a field of conduction velocity tensors, caused by the symmetries that are inherent to fully connected neural networks. To address this issue, we propose to quantify the uncertainty of the conduction velocity tensor estimation. To achieve this task, we use randomized prior neural networks \cite{Osband2018}. Here, an ensemble of $S_e$ neural networks is trained in parallel. Each of the ensemble members includes a ``prior'' network $\epsilon(\vec{x};\vec{\theta}_p)$ that shares the same architecture as the original network, but its parameters $\vec{\theta}_p$ are randomly initialized and then fixed, thus not updated during training. 
Using this approach, we define each $k$ model of the ensemble, as follows:
\begin{align}
    \hat d^k(\vec{x})&\coloneqq NN(\vec{x}; \vec{\theta}^k_D) + \lambda_p \epsilon_{d}^k(\vec{x}),&\label{eq:erpnn_d}\\
    \hat \phi_i^k(\vec{x})&\coloneqq NN(\vec{x}; \vec{\theta}_{\phi^k_i})+ \lambda_p \epsilon_{\phi_i}^k(\vec{x})& i\in[1,N]\label{eq:erpnn_phi},
\end{align}
where $\lambda_p$ is a hyperparameter that indicates the importance of the prior neural network. The parameters of each $k$ model (both the fixed and trainable part) are initialized independently using Glorot initialization \cite{glorot2010understanding}.

We redefine the loss function to aggregate the loss of each model as follows, 
\begin{equation}
    \mathcal{L}_{\text{ensemble}}\coloneqq\frac{1}{S_e}\sum_{k=1}^{S_e} \mathcal{L}(\vec{\theta}_D^k , \vec{\theta}_{\phi}^k),
\end{equation}
and optimize all the neural networks in parallel.  

\subsection{Fiber selection}
Since we have an ensemble of optimized models, we obtain a number of $S_e$ possible fiber orientations for a point on the geometry. If we need to determine what is the most appropriate fiber orientation to perform a simulation or another kind of procedure, we are required to choose only one fiber orientation per point. As such, we studied two vector selection approaches. Once a vector is selected, a notion of uncertainty is presented trough the disparity of the fiber with respect to the fibers on the ensemble.

\begin{figure}[ht]
    \centering
    \includegraphics[width=0.3\linewidth]{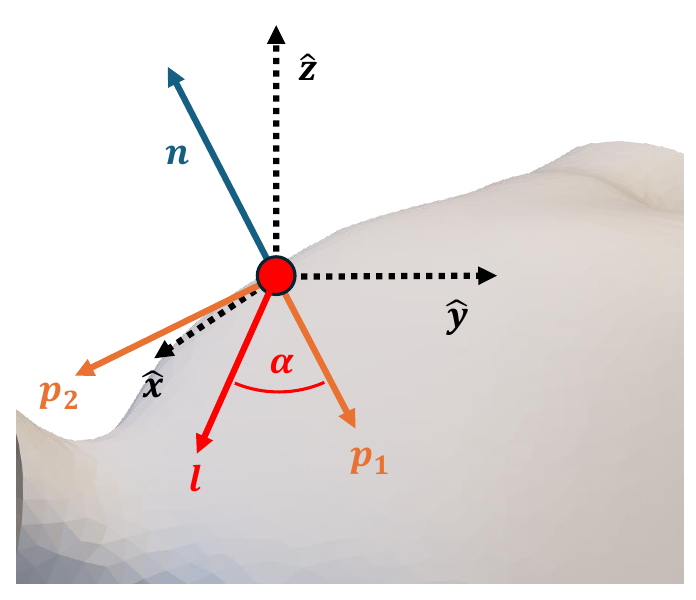}
    \includegraphics[width=0.38\linewidth]{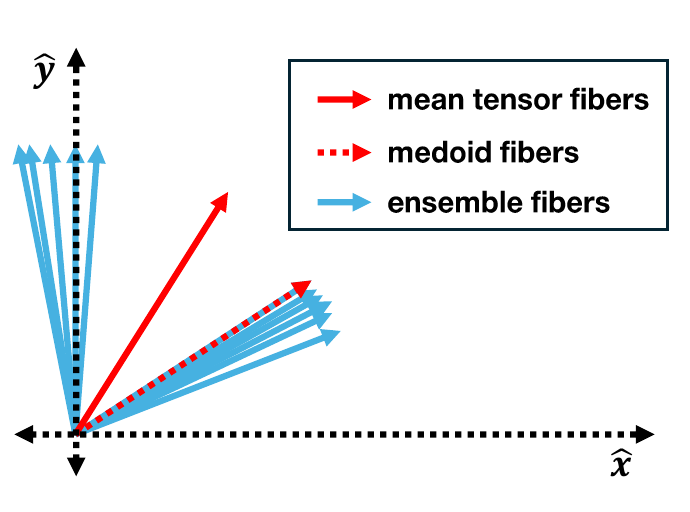}
    \caption{Left: Schematic of the fiber angle $\alpha(\vec{x})$ with respect to the orthonormal basis $\mathcal{B}(\vec{x})$ indicating. Right: Diagram of different behaviour of the fibers selected by Mean Tensor and Medoid approaches.}
    \label{fig:medoidmean}
\end{figure}

\subsubsection{Mean Tensor}
The first method corresponds to the exponential-log-mean tensor \cite{arsigny2005}, in further sections simply called Mean Tensor. This is a classical way to aggregate symmetric positive definite tensors, ensuring that the result is also a symmetric positive definite tensor. To calculate it, we first compute the rotation matrix for each $k$-model:
\begin{equation}
    \ten{R}_k(\vec{x}) =  \begin{bmatrix}
    a^k(\vec{x}) &  -\sqrt{1-a^k(\vec{x})^2}\\
    \sqrt{1-a^k(\vec{x})^2}& a^k(\vec{x}) 
  \end{bmatrix},
\end{equation}
then we take the mean of the rotated $\log$-scaled velocities,
\begin{equation}
    \ten{L}(\vec{x}) = \frac{1}{S_e} \sum_{k=1}^{S_e} 
 \ten{R}_k(\vec{x})
  \begin{bmatrix}
    \ln\left(e^k_1(\vec{x})\right) &  0\\
    0& \ln\left(e^k_2(\vec{x})\right)
  \end{bmatrix}\ten{R}_k(\vec{x})^T,
\end{equation}
and finally, we take the matrix exponential of the mean-logarithmic tensor,

\begin{equation}
    \ten{M}(\vec{x}) = e^{\boldsymbol{L}(\boldsymbol{x})}.
\end{equation}
Then, the fiber orientation $\vec{f}(\vec{x})$ corresponds to the eigenvector with the largest eigenvalue of $\ten{M}(\vec{x})$.

\subsubsection{Medoid}
The second method chooses a fiber orientation from the set of possible fibers given by the ensemble members, $\mathcal{F}(\vec{x}) = \{\vec{f}^1(\vec{x}), \vec{f}^2(\vec{x}), ...,\vec{f}^{S_e}(\vec{x})\}$. This allows us to enforce that the fiber chosen is an option given by the ensemble and is not a fiber outside of it, as it could be the case with the Mean Tensor method, an illustration is given on Fig.~\ref{fig:medoidmean}.

In particular, we defined the following notion of distance:
\begin{equation}
    d(\vec{v}_1, \vec{v}_2) = \left(1-|\vec{v}_1 \cdot \vec{v}_2|\right)^2,
\end{equation}
with $\vec{v}_1$ and $\vec{v}_2$ normalized vectors. The dot product between these vectors is the cosine of the angle between them. The result of this dot product will be 0 if the vectors are orthogonal and $1$ or $-1$ if they are parallel. The sign of the dot product reflects if the vectors point in the same direction ($1$) or not ($-1$). As the direction is not of interest, we take the absolute value of the dot product. Then we subtract it to $1$. In this way, $d(\vec{v}_1, \vec{v}_2)=0$ means that the vectors have the same orientation, and that $d(\vec{v}_1, \vec{v}_2)$ is always positive if $\vec{v}_1 \neq \vec{v}_2$. Then, $\forall \vec{v}_1, \vec{v}_2$ we have that $d(\vec{v}_1, \vec{v}_2)\leq0$. 
Then the representative fiber orientation is chosen, solving the following optimization problem:
\begin{equation}
    \vec{f}^*(\vec{x}) = \argmin_{\vec{f}(\vec{x})\in \mathcal{F}(\vec{x})} \median\left([d\left(\vec{f}(\vec{x}), \vec{f}^1(\vec{x})\right), d\left(\vec{f}(\vec{x}), \vec{f}^2(\vec{x})\right), ..., d\left(\vec{f}(\vec{x}), \vec{f}^{S_e}(\vec{x})\right)]\right)
\end{equation}

This approach, closely resembles the $k$-Medoids clustering algorithm \cite{Jin2010}, and as such, we refer to it as Medoid in following sections.

\subsubsection{Disparity}
As a metric of uncertainty in the fiber orientation field, we define the disparity between the selected fiber orientation from the ensemble ($\hat f^*$) and the fibers on the ensemble, as:
\begin{equation}
    \text{disparity}(\vec{x})=\arccos{\left(\frac{1}{S_e}\sum_{k=1}^{S_e}|\vec{\hat f}^*(\vec{x}) \cdot \vec{\hat f}^k(\vec{x})|\right)}. 
\end{equation}
This metric is closely related to the average orientation deviation from the selected fiber orientation. As we are taking the mean of the absolute value of the cosine of the angles between the selected fiber and the fibers on the ensemble. If all ensemble members have similar orientations to the selected fibers, this metric should be close to zero, while if models disagree in orientation, this quantity will increase.

\section{Numerical assessment}
\label{sec:numerical}
Fibernet and $\Delta$-Fibernet were implemented  using Google's JAX library \cite{jax2018github} and are available on \url{https://github.com/fsahli/Delta-fibernet}. For the training, we used 50.000 iterations of the ADAM optimizer \cite{Kingma2017} with its default hyperparameters. On each iteration we passed the full set of measured activation maps to evaluate $\mathcal{L}_{\text{data}}$, and a random batch of points was selected as collocation points to evaluate $\mathcal{L}_{\text{eiko}}$, $\mathcal{L}_{\text{CV}}$ and $\mathcal{L}_{\text{ang}}$. The size of the batch was 64 for all cases. The same architecture was also used for all cases, 5 hidden layers of 20 neurons for the activations times, and 7 hidden layers of 20 neurons for the conductivity tensor. Furthermore, the following hyperparameters were also fixed for all cases $\{S_e:20, \lambda_{p}:10^{-3}\}$. For $\Delta$-Fibernet the 10 first eigenfunctions were used.

We tested our methodology with 2 experiments, the case of a synthetic atrial geometry with rule-based assigned fibers, and 7 different ex-vivo atrial geometries with DT Imaging measured fibers. For all cases, we simulated 3 activation maps as ground-truth with a Fast Iterative solver for the eikonal equation~\cite{Grandits2021}, selecting a random initial point for the first map, and choosing the farthest point for the second, then choosing the farthest points with respect to the 2 existing points as third map. We added noise to these simulations, following a normal distribution scaled by the level of noise. Finally, we randomly select $N_i=\lfloor A \rho \rfloor$ points for each activation map, where $A$ is the area of the geometry in $\text{cm}^2$ and $\rho$ is the sampling density in $\text{points}/\text{cm}^2$. 

For the activation maps, we report the Root-Mean-Squared Error ($\text{RMSE}$), defined as,
\begin{equation}
    \text{RMSE}(\phi_i, \hat \phi_i) = \sqrt{\frac{1}{N_p}\sum_{\boldsymbol{x}\in \mathcal{S}} \left(\phi_i(\vec{x})-\hat \phi_i(\vec{x})\right)^2},
\end{equation}
with $N_p$ being the number of points on the mesh that describes $\mathcal{S}$, this is calculated for each $i$-map. For the fiber orientation, we report the angle between the ground truth fiber vs. the predicted fiber, which is defined as,
\begin{equation}
    \beta(\vec{x}) = \arccos{|\vec{f}(\vec{x}) \cdot  \vec{\hat f}(\vec{x})|}.
\end{equation}
Using $\beta(x)$, we report a global fiber orientation error (FE), defined as:
\begin{equation}
\text{FE} = \median_{\boldsymbol{x}\in \mathcal{S}} \left(\beta(\vec{x})\right).
\end{equation}

Lastly, using these metrics and the first of the geometries with DT Imaging measured fibers, we performed a grid-search to select the optimal hyperparameters for Fibernet and $\Delta$-Fibernet. We fixed the density $16$ $[\text{points}/{\text{cm}^2}]$ and the noise level to $2$ [ms]. The resulting hyperparameters, otherwise specified, were used on all geometries (including the synthetic atrial geometry), and corresponds to for Fibernet $\{\lambda_{\text{data}}:10^{-1}, \lambda_{\text{eiko}}:10^{-3}, \lambda_{\text{CV}}:10^{-5}, \lambda_{\text{ang}}:10^{-9}\}$, and for $\Delta$-Fibernet $\{\lambda_{\text{data}}:10^{-1}, \lambda_{\text{eiko}}:10^{-4}, \lambda_{\text{CV}}:10^{-5}, \lambda_{\text{ang}}:10^{-8}\}$.

\subsection{Atrial geometry with rule-based fibers}
\begin{figure}[h!tbp]
    \centering
    \includegraphics[width=0.8\linewidth]{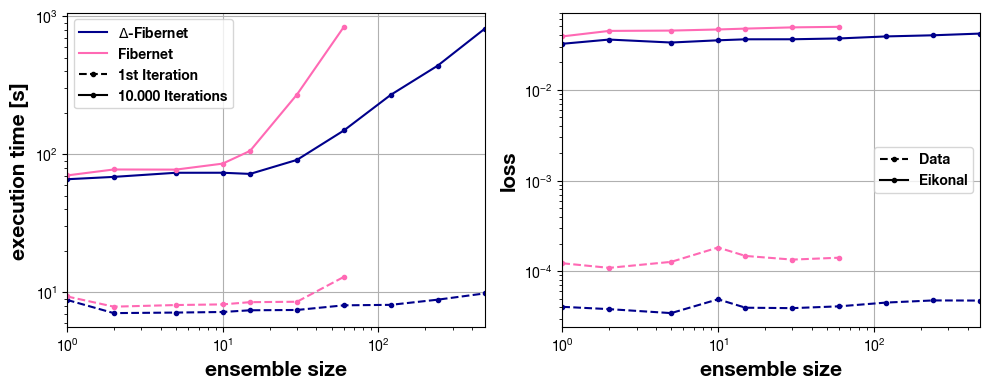}
    \caption{Effects of ensemble size on execution time and loss minimization, for $\Delta$-Fibernet, in blue, and Fibernet, in magenta. The left graph presents the execution time for the first iteration and for 10.000 following iterations. In the right, the resulting $\mathcal{L}_{\text{data}}$ and $\mathcal{L}_{\text{eiko}}$ after 10.001 iterations is reported.}
    \label{fig:ensemblesize}
\end{figure}

We use a synthetic left atrium geometry of the endocardial wall with rule-based fibers to compare the proposed $\Delta$-Fibernet against Fibernet. 
First, we study the performance of our implementation for different ensemble sizes, monitoring the execution time and the ensemble loss. We use a laptop with a 12th Gen Intel(R) Core(TM) i7-12700H CPU, an NVIDIA GeForce RTX 3070 Laptop GPU (8 GB VRAM) and 64 GB of RAM. The activation maps used had a density of $8$ $[\text{points}/\text{cm}^2]$ and no noise was added, for the training the hyperparameters were fixed as $\{\lambda_{\text{data}}:1, \lambda_{\text{eiko}}:10^{-4}, \lambda_{\text{CV}}:10^{-5}, \lambda_{\text{ang}}:10^{-8}\}$ for both implementations. 
The time that it takes for the first iteration includes the compilation of the code, thus is different to the following iterations. We report the time that it takes to run the first iteration and the following 10.000 iterations. For the losses, we report $\mathcal{L}_{\text{data}}$ and $\mathcal{L}_{\text{eiko}}$ at the 10.001 iteration. The results are shown on Fig. \ref{fig:ensemblesize}. 

As it can be seen on Fig. \ref{fig:ensemblesize}, the ensemble size does affect the execution time for small ensemble sizes. The time to execute 10,000 iterations after an ensemble size of 10 starts to grow exponentially, with the Fibernet implementation growing faster, being closer to 1000 seconds at an ensemble size of 60. Meanwhile, $\Delta$-Fibernet is closer to 1000 seconds at an ensemble size of 480. Ignoring the results for ensemble size of 1, it is noted that the first iteration grows at a regular pace in the case of $\Delta$-Fibernet. The same is true for Fibernet until an ensemble size of 30, where it can be seen how it stars to exponentially grow. These differences can be explained by the fact that $\Delta$-Fibernet is not using automatic differentiation. Therefore, the number of calls made to the function is lower and thus it can run faster. Still, the execution time will grow with the ensemble size, as the number of parameters to optimize grows linearly with it.
Looking at the losses, both grow at a slow rate as the ensemble grows. While the losses for Fibernet are greater than to those of $\Delta$-Fibernet, the difference is more visible for $\mathcal{L}_{\text{data}}$. It can be seen that randomized prior functions are an efficient alternative to estimate uncertainty, with little overhead when compared to training a single neural network.

\begin{figure}
    \centering
    \includegraphics[width=0.9\linewidth]{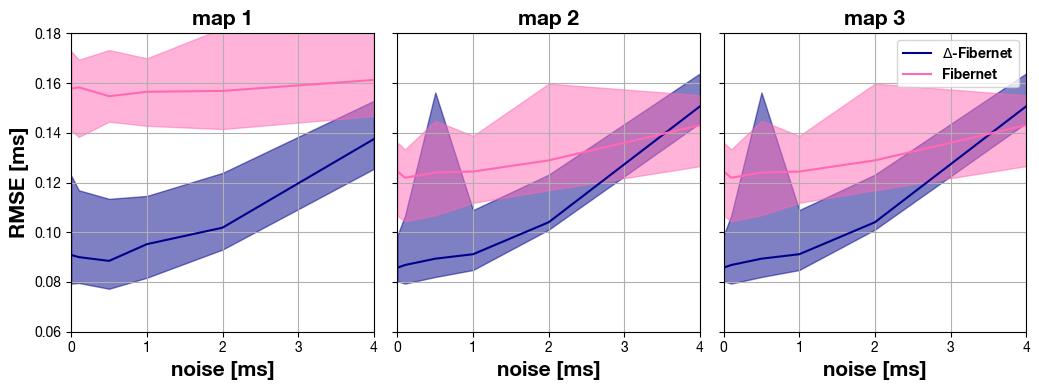}
    \caption{Effect of noise on the prediction of the activation maps for both, $\Delta$-Fibernet and Fibernet. The solid line present the median result, with the filled area presenting the area between the best case and the worst.}
    \label{fig:fullresults}
\end{figure}

\begin{figure}
    \centering
    \includegraphics[width=0.49\linewidth]{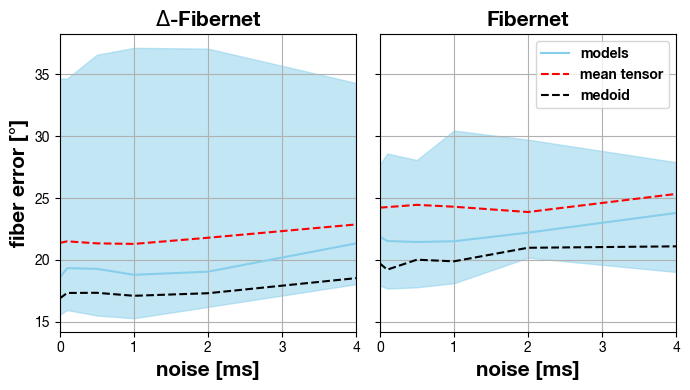}
    \includegraphics[width=0.49\linewidth]{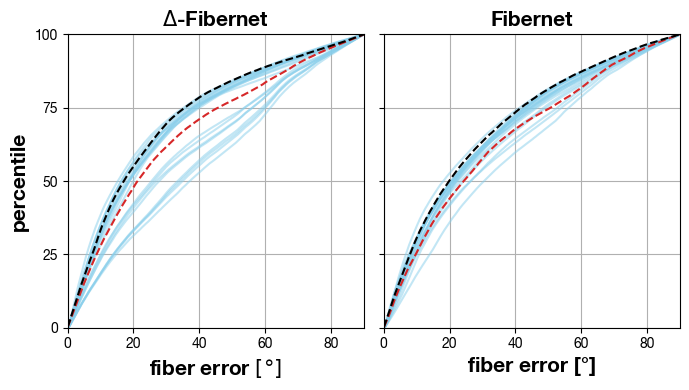}
    \caption{The first and second columns, present the effect of noise on the prediction of the orientation fibers maps for both, $\Delta$-Fibernet and Fibernet respectively. The solid line present the median result, with the filled area presenting the area between the best case and the worst. Meanwhile, on the third and fourth columns, the cumulative error for each model in the ensemble for $\Delta$-Fibernet and Fibernet respectively, for the case with 1 [ms] of noise. For all columns, the red dotted line presents the Mean Tensor fibers and the black dotted line the Medoid fibers.}
    \label{fig:onecase}
\end{figure}

\begin{figure}[t]
    \centering
    \includegraphics[width=\linewidth]{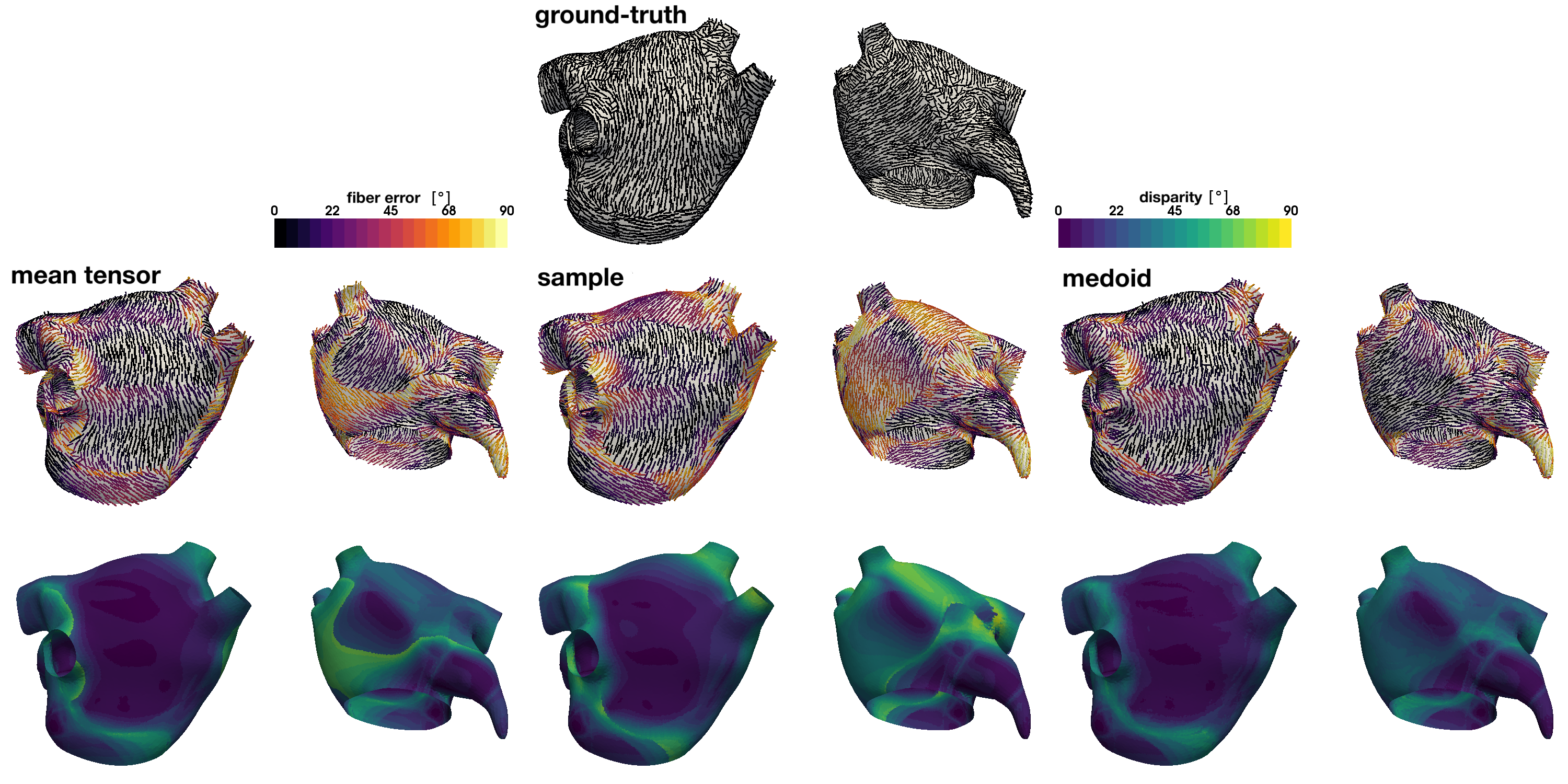}
    \caption{Predicted fibers using $\Delta$-Fibernet for the case of maps with 1 [ms] of noise. The Mean Tensor fibers are shown on the left, a sampled model from the ensemble, in the middle, and the Medoid fibers at the right. The Fiber error is shown on the second row, and the disparity on the third. The first row shows the ground-truth fibers.}
    \label{fig:simple3dresults}
\end{figure}

Following the ensemble size study, we test the robustness to noise of both methods, fixing a density of $16$ $[\text{points}/{\text{cm}^2}]$ and using the hyperparameters from the grid-search exploration. The results are displayed in Fig.~\ref{fig:fullresults}. We also show the case where the noised is fixed at 1 [ms] in Fig.~\ref{fig:onecase} as a cumulative error for each model in the ensemble and for Fibernet and $\Delta$-Fibernet. Lastly, in Fig.~\ref{fig:simple3dresults}, we show the geometry with its ground-truth fibers and the fibers obtained with $\Delta$-Fibernet, comparing the Mean Tensor, a sample from the ensemble and the Medoid.  

On Fig.~\ref{fig:fullresults} it can be seen how $\Delta$-Fibernet learns the activation maps more accurately when the noise is lower than 3 [ms], and that after this Fibernet has lower errors. Meanwhile, in both cases, the error grows directly with the noise. But $\Delta$-Fibernet has the steepest growth. This could be explained as the use of eigenfunctions of increasing frequency to encode the surface into the neural network. This method allows us to learn high frequency and low frequency signals at the same time. The range of the frequencies being controlled by the number of eigenfunctions we select. Therefore, $\Delta$-Fibernet can also learn the high frequency components. Fibernet, on the other hand, is based on a multilayer perceptron, which is known to suffer from a frequency bias, where low frequencies are learned first \cite{luo2019theory,molina2024understanding}. Thus, Fibernet has less capacity to learn the high frequency component of the noise.

On the other hand, the fiber orientation error on Fig.~\ref{fig:onecase}, has less dependency on the level of noise and both methods share a similar trend. With the median of the error of the models in the ensemble of Fibernet being higher than those of $\Delta$-Fibernet. However, $\Delta$-Fibernet has a larger range of error than Fibernet. This means that the worst result of $\Delta$-Fibernet is worse than the worst of Fibernet, and that the best of $\Delta$-Fibernet, is better than the one of Fibernet.  Overall, the Medoid fiber orientation perform better than the Mean Tensor fiber orientation. The Medoid fiber orientation of $\Delta$-Fibernet has the best results with a $3\degree$ difference with respect to the Medoid fiber orientation for Fibernet. The less pronounced growth with respect to the noise could be an effect of the regularization used for the conductivity tensor neural network. Not only we explicitly use the Huber regularization of its outputs, but its result depends on all $N$ maps simultaneously. This effectively lowers the effect of noise for the conductivity tensor.

Looking into the cumulative error reported on Fig.~\ref{fig:onecase}, $\Delta$-Fibernet has better overall cumulative errors than Fibernet. It can be observed that for the models on $\Delta$-Fibernet two general behaviors are formed, one with better performance than the other. This could explain the greater range of the $\Delta$-Fibernet results with respect to Fibernet's, where only one clear behavior is formed. On the other hand, it can be seen how the Mean Tensor fiber orientations are greatly affected by outliers. As for $\Delta$-Fibernet the Mean Tensor curve separates the two behaviors. For Fibernet, the Mean Tensor fiber orientation curve is closer to the models with the worst performance, even though they are few. It can be seen that the Medoid fiber orientation tends to resemble the best performing model on the ensemble for both methods.

In Fig.~\ref{fig:simple3dresults}, we compare the predicted fiber orientation of a model of the ensemble (sample), the Mean Tensor fiber orientation and the Medoid fiber orientation. Here, it can be seen how a model in the ensemble can have high error regions that are learned better by the Medoid and the Mean Tensor. The Mean Tensor fiber orientation has higher levels of disparity, meaning that the fibers obtained trough this method may not be adequate to describe the results of the ensemble. As expected, the Medoid fiber orientation has the lowest disparity in this case. By definition, the Medoid method selects the fiber orientation that minimizes the overall distance to the others in the ensemble, which will end up minimizing the disparity. 

\subsection{Atrial geometries with diffusion tensor fibers}
\begin{figure}[t]
    \centering
    \includegraphics[width=1\linewidth]{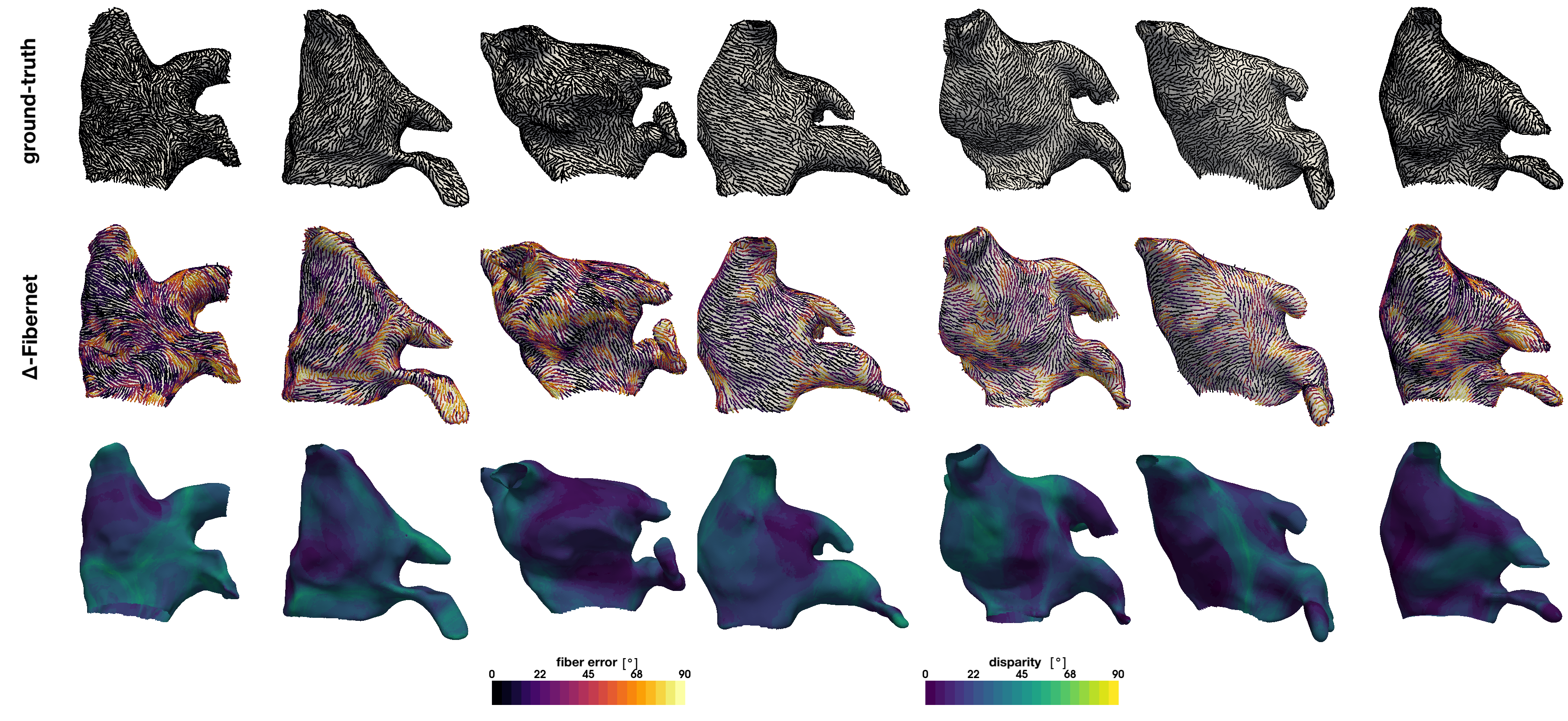}
    \caption{Predicted fibers using $\Delta$-Fibernet for the case of maps with 1 [ms] of noise for the geometries with DT Imaging assigned fibers. Only the Medoid fibers are shown. At the top, the ground-truth fibers are shown, in the middle row, the fiber's errors, and at the bottom the disparity. The anteroposterior view is shown.}
    \label{fig:7geometrie}
\end{figure}

For the next experiment, we consider 7 different left atrium geometries with ex-vivo measured fibers representing the endocardial wall \cite{Pashakhanloo2016geometries}, upper row of Fig. \ref{fig:7geometrie}

\begin{figure}[h!tpb]
    \centering
    \includegraphics[width=0.8\linewidth]{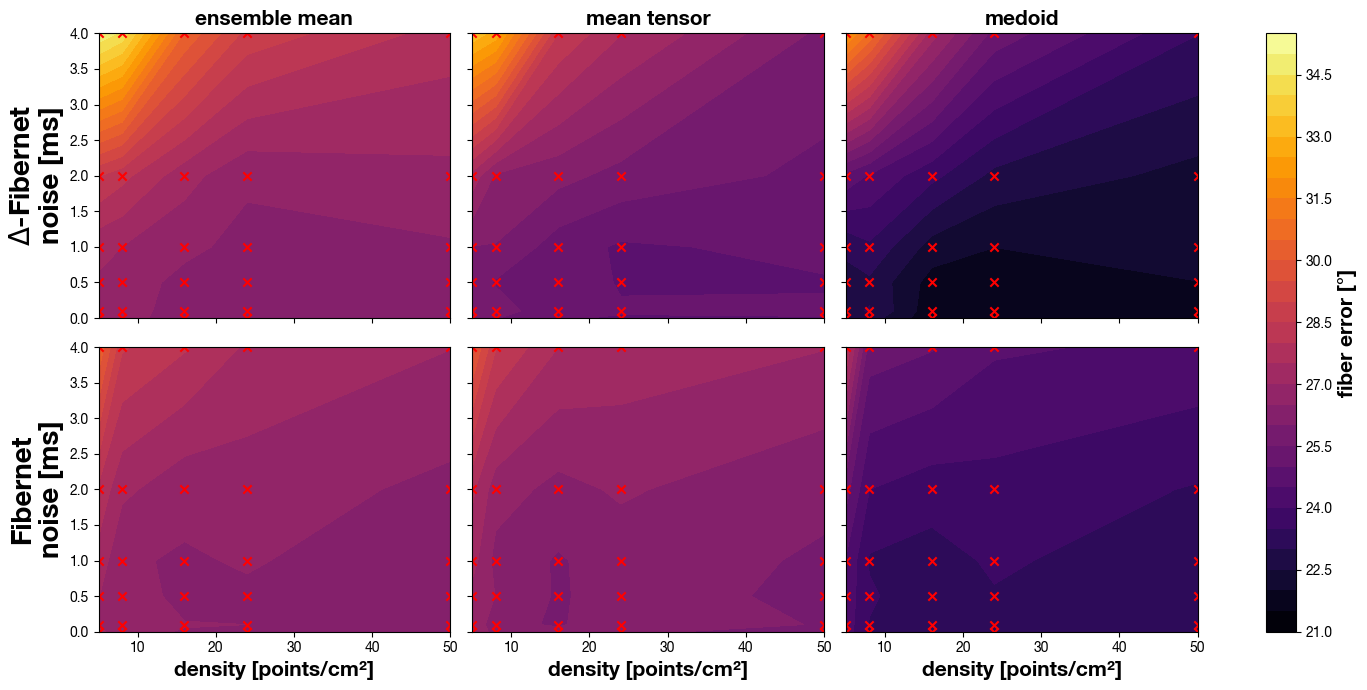}
    \caption{Effect of noise and density on the fiber orientaiton error for all the geometries with DTI fibers. The mean over all geometries is reported. On the first column, the mean fiber orientation error for all members of the ensemble, on the second, for the Mean Tensor, and lastly, for the Medoid. The upper row presents $\Delta$-Fibernet, and the bottom, Fibernet. The red crosses represent the tested combinations.}
    \label{fig:contourgeometry}
\end{figure}

To check the effect of noise and data density on these geometries, we tested 30 combinations of these variables, the mean of the fiber error on the 7 geometries for each combination of variables and separated into $\Delta$-Fibernet and Fibernet, and ensemble, Mean Tensor and Medoid fibers, is reported on Fig.~\ref{fig:contourgeometry}. At first glance, the effect of having a lower density and greater noise, left upper corner, can be noticed with the fiber orientation error increasing for both Fibernet and $\Delta$-Fibernet. Nevertheless, this effect is accentuated on the $\Delta$-Fibernet implementation. This is an expected result, as it was discussed before, $\Delta$-Fibernet's learning process is inclined to learn the high frequency components of the noise. Therefore, in the cases with high noise and small data density, the models learn noisy activation maps, that in effect increase the fiber orientation error. This effect is diminished as the density grows, making the difference on error between a case without noise and one with $4$ [ms] of noise, of $1\degree$  and $2\degree$  for the Fibernet and $\Delta$-Fibernet respectively, when the density is grater than $24$ $[\text{points}/{\text{cm}^2}]$. Another insight that can be appreciated from the figure, is that the fiber orientation selected by the Medoid method (right column) has a lower error as the fiber selected using the Mean Tensor approach (middle column), and to the mean of the results in the ensembles (first column). Using the Medoid, reduces the error between $3 - 4\degree$ with respect to the other methods. Following this, it is also of note that $\Delta$-Fibernet has a minimum error with $21.6\degree$, compared to the best result of Fibernet, which corresponds to $23.1\degree$. These results are consistent with the ones presented for the synthetic atrial geometry in the previous section.

Furthermore, the results for $\Delta$-Fibernet with a density of $16$ $[\text{points}/{\text{cm}^2}]$ and a noise of 1 [ms] are summarized in Table \ref{tab:7geometries}. The Medoid fiber orientation for each geometry is shown on Fig.~\ref{fig:7geometrie}. Looking at the table, the activation maps error are close to the expected value of 1 [ms] for the noise level of 1 [ms]. If the error is lower than 1 [ms], the models are learning to denoise the activation maps. This is the case for most geometries, with case number 5 being the exception. As for the fiber orientaiton errors, the worst case corresponds to the fourth geometry, with the mean error in the ensemble of $30.8\degree$. However, this error is greatly reduced by considering the Medoid, which is almost as good as the best performing models for this case. This trend is extended to all cases, where the Medoid fiber orientation is better than the Mean Tensor fiber, with a margin varying in between $1-4\degree$. Meanwhile, the Mean Tensor fibers are closer to the mean error of the ensemble, with only on the cases of the first and second geometries being lower for a considerable margin of $3\degree$.

Finally, it is remarkable that the disparity closely follows the fiber error. While the latter is not known in real case scenarios, the former is computed from the ensemble and always available. Therefore, the disparity is a good measure of the local reconstruction error.

\begin{table}[h]
\centering
\caption{Summary of the predicted activation maps and fibers, for $\Delta$-Fibernet for the case with 1 [ms] of noise.}
\label{tab:7geometries}
\begin{tabular}{l|ccc|cccc|}
  & \multicolumn{3}{c|}{Activation maps error [$\mu \text{s}$]} & \multicolumn{4}{c|}{Fibers error [$\degree$ ]}                                                                                                         \\
  \# & $\phi_1$        & $\phi_2$        & $\phi_3$        & \begin{tabular}[c]{@{}c@{}}Ensemble \\ Mean\end{tabular} & \multicolumn{1}{c|}{Range}         & \begin{tabular}[c]{@{}c@{}}Mean \\ Tensor\end{tabular} & Medoid                         \\ \hline
1 & $79 \pm 5$& $80 \pm 6$& $76 \pm 5$& $27.2$ & \multicolumn{1}{c|}{$21.7 - 41.5$} & $23.8$                                                 & $\mathbf{22.4}$\\
2 & $95 \pm 8$& $110 \pm 11$& $92 \pm 6$& $28.3$ & \multicolumn{1}{c|}{$22.8 - 39.8$} & $25.4$                                                 & $\mathbf{23.4}$ \\
3 & $98 \pm 7$& $102 \pm 6$& $98 \pm 8$& $26.5$ & \multicolumn{1}{c|}{$22.3 - 41.4$} & $26.2$                                                 & $\mathbf{22.7}$ \\
4 & $90 \pm 7$& $100 \pm 8$& $91 \pm 7$& $30.8$ & \multicolumn{1}{c|}{$24.6 - 45.1$} & $28.6$                                                 & $\mathbf{25.8}$ \\
5 & $117 \pm 8$& $111 \pm 8$& $116 \pm 4$& $29.3$ & \multicolumn{1}{c|}{$25.5 - 39.5$} & $28.5$                                                 & $\mathbf{25.0}$ \\
6 & $97 \pm 7$& $105 \pm 7$& $94 \pm 10$& $22.1$ & \multicolumn{1}{c|}{$18.7 - 28.3$} & $21.7$                                                 & $\mathbf{18.4}$ \\
7 & $84 \pm 3$& $91 \pm 10$& $92 \pm 7$& $22.3$ & \multicolumn{1}{c|}{$18.6 - 28.9$} & $22.5$                                                 & $\mathbf{18.4}$ \\ \hline
\end{tabular}
\end{table}

\section{Discussion}
In this work, we presented a novel method to estimate the atrial fiber orientation from electroanatomical maps with physics-informed neural networks. We have expanded our previous methodology with 3 key developments: first $\Delta$-Fibernet allow us to learn directly in a representation of the surface, improving the fiber orientation inference with respect to traditional physics-informed neural networks. Second, we can now provide uncertainty estimates of conduction velocity tensor with little computational overhead. Finally, we have devised a novel way to combine the results from multiple ensemble members, which is always close to the best performing member in the form of the Medoid. 

To validate the proposed developments, we worked with 8 left atrium geometries, a synthetic and simpler geometry, and 7 real geometries with DT Imaging measured fibers. We tested how the ensemble size affects the training time, showing that dropping the automatic differentiation on Fibernet for the element differentiation on $\Delta$-Fibernet helps to reduce the computational cost of training without a loss of accuracy. Furthermore, we note that training $\Delta$-Fibernet only takes around 7 minutes to train with modest hardware. In this regard, these are times that could allow a clinical implementation of this method. The computational cost could be further reduced by improving the hardware and also including prior information about the fiber orientation. We also characterize the effect of the noise with respect to the density of the measurements in 7 fiber orientations derived from DT Imaging, with $\Delta$-Fibernet being more sensible to noise than the original Fibernet when the noise surpasses 2 [ms]. However, increasing the density of the measurements have shown to improve the results in fiber orientation for both Fibernet and $\Delta$-Fibernet. When the noise was fixed at 1 [ms] and the density to 16 $[\text{points}/{\text{cm}^2}]$, the $\Delta$-Fibernet Medoid selected fibers have shown to perform consistently better than those selected by the Mean Tensor approach, and also from Fibernet. 

Despite the promising results, our work presents some limitations. We only analyzed the fiber inference on the left atrium, under the assumption that they are thin enough to be represented by a surface. We could directly apply this method to the right atrium, but applying to the ventricles would require a reformulation. Reducing the atria to a surface implies that we can only recover the fiber orientation where the activation map was measured. Another limitation is that we only consider synthetic activation maps, which were corrupted with noise. Since there is no method to quantify the atrial fiber orientation \textit{in vivo}, it is not possible to use electroanatomical maps to acquire the activations maps and construct a complete dataset. The complete validation of $\Delta$-Fibernet would require animal experiments, to acquire the activation maps \textit{in vivo} and then obtain the fiber orientation from DT Imaging \textit{ex vivo}, which we plan to pursue in the future. Related to the previous point, we require 3 different activations maps, that even though is possible to acquire during an electroanatomical mapping procedure, it is uncommon in clinical practice. We have demonstrated that at least 3 maps are required to reconstruct the conduction velocity tensor \cite{Ruiz2022}. It would be possible to relax this condition in 2 different ways. First, fixing some parameters of this tensor, such as the anisotropy ratio and/or the conduction velocities, should enable the recovery of the fiber orientation from fewer maps. The second option is introducing a strong prior on the fiber orientation, which could be derived from rule-based algorithms \cite{Piersanti2021,piersanti2024defining}. These reduced cases could still enhance the personalization of cardiac digital twins, opening the door to use non-invasive measurement techniques, such as the electrocardiogram \cite{alvarez2023probabilistic,Pezzuto2021ECG,GeodesicBP2024}.

Overall, we have developed a robust algorithm for the identification of the conduction velocity tensor from sparse activation maps. We hope that this work would enable faster and more precise personalization of cardiac models to enable better medical care for cardiac arrhythmias.


\bibliographystyle{elsarticle-harv}
\bibliography{litra}

\section{Competing Interests}

None of the authors has any conflicts of interest

\section{Funding}

FSC and EM acknowledge the support of the project ERAPERMED-134 from ANID. This work was also funded by ANID – Millennium Science Initiative Program – ICN2021\_004 and National Center for Artificial Intelligence CENIA FB210017, Basal ANID to FSC. EM also acknowledge ANID BECAS/DOCTORADO NACIONAL 21240538. SP acknowledges the support of the CSCS-Swiss National Supercomputing Centre project no.~s1074 and the PRIN-PNRR project no.~P2022N5ZNP.

\end{document}